# Obtaining Calibrated Probabilities from Boosting

**Alexandru Niculescu-Mizil**
Department of Computer Science
Cornell University, Ithaca, NY 14853
alexn@cs.cornell.edu

**Rich Caruana**
Department of Computer Science
Cornell University, Ithaca, NY 14853
caruana@cs.cornell.edu

## Abstract

Boosted decision trees typically yield good accuracy, precision, and ROC area. However, because the outputs from boosting are not well calibrated posterior probabilities, boosting yields poor squared error and cross-entropy. We empirically demonstrate why AdaBoost predicts distorted probabilities and examine three calibration methods for correcting this distortion: Platt Scaling, Isotonic Regression, and Logistic Correction. We also experiment with boosting using log-loss instead of the usual exponential loss. Experiments show that Logistic Correction and boosting with log-loss work well when boosting weak models such as decision stumps, but yield poor performance when boosting more complex models such as full decision trees. Platt Scaling and Isotonic Regression, however, significantly improve the probabilities predicted by both boosted stumps and boosted trees. After calibration, boosted full decision trees predict better probabilities than other learning methods such as SVMs, neural nets, bagged decision trees, and KNNs, even after these methods are calibrated.

## 1 Introduction

In a recent evaluation of learning algorithms [Caruana and Niculescu-Mizil, 2005], boosted decision trees had excellent performance on metrics such as accuracy, lift, area under the ROC curve, average precision, and precision/recall break even point. However, boosted decision trees had poor squared error and cross-entropy because AdaBoost does not produce good probability estimates.

Friedman, Hastie, and Tibshirani [2000] provide an explanation for why boosting makes poorly calibrated predictions. They show that boosting can be viewed as an additive logistic regression model. A consequence of this is that the predictions made by boosting are trying to fit a logit of the true probabilities, as opposed to the true probabilities themselves. To get back the probabilities, the logit transformation must be inverted.

In their treatment of boosting as a large margin classifier, Schapire et al. [1998] observed that in order to obtain large margin on cases close to the decision surface, AdaBoost will sacrifice the margin of the easier cases. This results in a shifting of the predicted values away from 0 and 1, hurting calibration. This shifting is also consistent with Breiman's interpretation of boosting as an *equalizer* (see Breiman's discussion in [Friedman *et al.*, 2000]). In Section 2 we demonstrate this probability shifting on real data.

To correct for boosting's poor calibration, we experiment with boosting with log-loss, and with three methods for calibrating the predictions made by boosted models to convert them to well-calibrated posterior probabilities. The three post-training calibration methods are:

**Logistic Correction:** a method based on Friedman et al.'s analysis of boosting as an additive model

**Platt Scaling:** the method used by Platt to transform SVM outputs from $[-\infty, +\infty]$ to posterior probabilities [1999]

**Isotonic Regression:** the method used by Zadrozny and Elkan to calibrate predictions from boosted naive Bayes, SVM, and decision tree models [2002; 2001]

Logistic Correction and Platt Scaling convert predictions to probabilities by transforming them with a sigmoid. With Logistic Correction, the sigmoid parameters are derived from Friedman et al.'s analysis. With Platt Scaling, the parameters are fitted to the data using gradient descent. Isotonic Regression is a general-purpose non-parametric calibration method that assumes probabilities are a monotonic transformation (not just sigmoid) of the predictions.

An alternative to training boosted models with AdaBoost and then correcting their outputs via post-training calibration is to use a variant of boosting that directly optimizes cross-entropy (log-loss). Collins, Schapire and Singer [2002] show that a boosting algorithm that opti-mizes log-loss can be obtained by simple modification to the AdaBoost algorithm. Collins et al. briefly evaluate this new algorithm on a synthetic data set, but acknowledge that a more thorough evaluation on real data sets is necessary.

Lebanon and Lafferty [2001] show that Logistic Correction applied to boosting with exponential loss should behave similarly to boosting with log-loss, and then demonstrate this by examining the performance of boosted stumps on a variety of data sets. Our results confirm their findings for boosted stumps, and show the same effect for boosted trees.

Our experiments show that boosting full decision trees usually yields better models than boosting weaker stumps. Unfortunately, our results also show that boosting to directly optimize log-loss, or applying Logistic Correction to models boosted with exponential loss, is only effective when boosting weak models such as stumps. Neither of these methods is effective when boosting full decision trees. Significantly better performance is obtained by boosting full decision trees with exponential loss, and then calibrating their predictions using either Platt Scaling or Isotonic Regression. Calibration with Platt Scaling or Isotonic Regression is so effective that after calibration boosted decision trees predict better probabilities than any other learning method we have compared them to, including neural nets, bagged trees, random forests, and calibrated SVMs.

In Section 2 we analyze the predictions from boosted trees from a qualitative point of view. We show that boosting distorts the probabilities in a consistent way, generating sigmoid-shaped reliability diagrams. This analysis motivates the use of a sigmoid to map predictions to well-calibrated probabilities. Section 3 describes the three calibration methods. Section 4 presents an empirical comparison of the three calibration methods and the log-loss version of boosting. Section 5 compares the performance of boosted trees and stumps to other learning methods.

## 2 Boosting and Calibration

In this section we empirically examine the relationship between boosting's predictions and posterior probabilities. One way to visualize this relationship when the true posterior probabilities are not known is through reliability diagrams [DeGroot and Fienberg, 1982]. To construct a reliability diagram, the prediction space is discretized into ten bins. Cases with predicted value between 0 and 0.1 are put in the first bin, between 0.1 and 0.2 in the second bin, etc. For each bin, the mean predicted value is plotted against the true fraction of positive cases in the bin. If the model is well calibrated the points will fall near the diagonal line.

The bottom row of Figure 1 shows reliability plots on a large test set after 1,4,8,32,128, and 1024 stages of boosting Bayesian smoothed decision trees [Buntine, 1992]. The top of the figure shows histograms of the predicted values for the same models. The histograms show that as the number of steps of boosting increases, the predicted values are pushed away from 0 and 1 and tend to collect on either side of the decision surface. This shift away from 0 and 1 hurts calibration and yields sigmoid-shaped reliability plots.

Figure 2 shows histograms and reliability diagrams for boosted decision trees after 1024 steps of boosting on eight test problems. (See Section 4 for more detail about these problems.) The figures present results measured on large independent test sets not used for training. For seven of the eight data sets the predicted values after boosting do not approach 0 or 1. The one exception is LETTER.P1, a highly skewed data set that has only 3% positive class. On this problem some of the predicted values do approach 0, though careful examination of the histogram shows that there is a sharp drop in the number of cases predicted to have probability near 0.

All the reliability plots in Figure 2 display sigmoid-shaped reliability diagrams, motivating the use of a sigmoid to map the predictions to calibrated probabilities. The functions fitted with Platt's method and Isotonic Regression are shown in the middle and bottom rows of the figure.

## 3 Calibration

In this section we describe three methods for calibrating predictions from AdaBoost: Logistic Correction, Platt Scaling, and Isotonic Regression.

### 3.1 Logistic Correction

Before describing Logistic Correction, it is useful to briefly review AdaBoost. Start with each example in the train set $(x_i, y_i)$ having equal weight. At each step $i$ a weak learner $h_i$ is trained on the weighted train set. The error of $h_i$ determines the model weight $\alpha_i$ and the future weight of each training example. There are two equivalent formulations. The first formulation, also used by Friedman, Hastie, and Tibshirani [2000] assumes $y_i \in \{-1, 1\}$ and $h_i \in \{-1, 1\}$. The output of the boosted model is:

$$F(x) = \sum_{i=1}^{T} \alpha_i h_i(x) \quad (1)$$

Friedman et al. show that AdaBoost builds an additive logistic regression model for minimizing $E(exp(-yF(x)))$. They show that $E(exp(-yF(x)))$ is minimized by:

$$F(x) = \frac{1}{2} log \frac{P(y=1|x)}{P(y=-1|x)} \quad (2)$$

This suggests applying a logistic correction in order to get back the conditional probability:

$$P(y=1|x) = \frac{1}{1 + exp(-2F(x))} \quad (3)$$

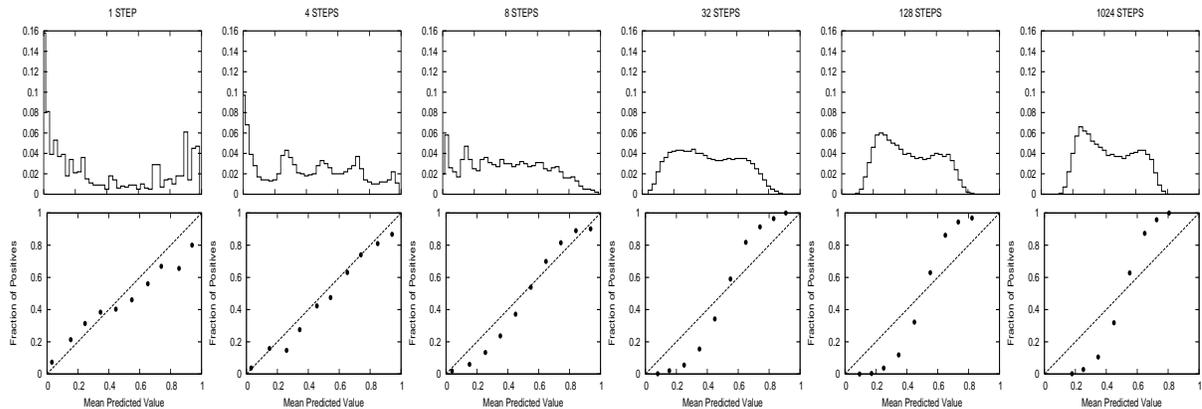

Figure 1: Effect of boosting on the predicted values. Histograms of the predicted values (top) and reliability diagrams (bottom) on the test set for boosted trees at different steps of boosting on the COV_TYPE problem.

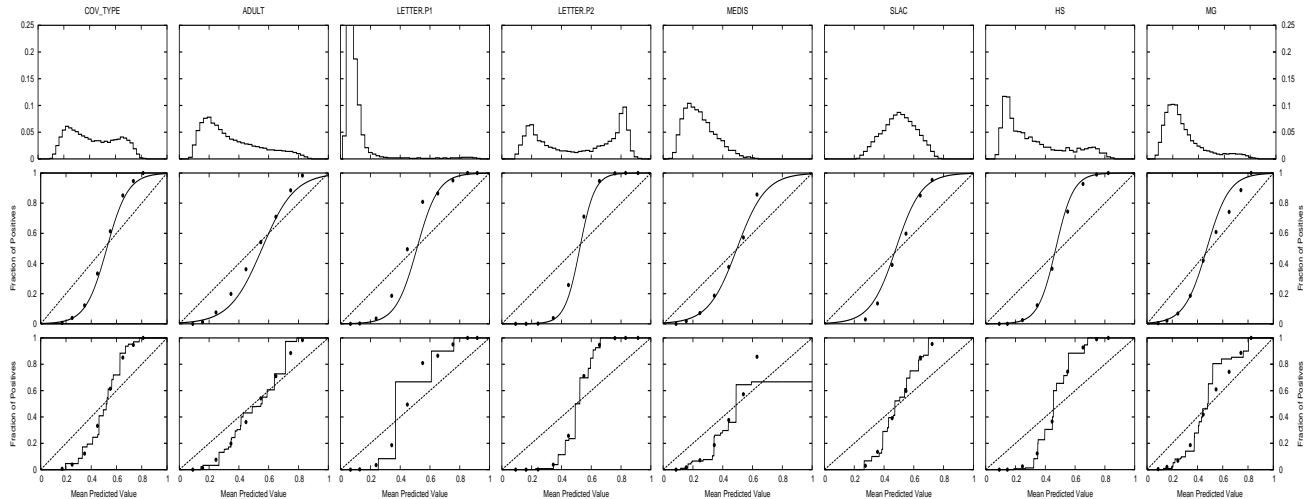

Figure 2: Histograms of predicted values and reliability diagrams for boosted decision trees.

As we will see in section 4, this logistic correction works well when boosting simple base learners such as decision stumps. However, if the base learners are powerful enough that the training data becomes fully separable, after correction the predictions will become only 0's and 1's [Rosset et al., 2004] and thus have poor calibration.

### 3.2 Platt Calibration

An equivalent formulation of AdaBoost assumes $y_i \in \{0, 1\}$ and $h_i \in \{0, 1\}$. The output of the boosted model is

$$f(x) = \frac{\sum_{i=1}^{T} \alpha_i h_i(x)}{\sum_{i=1}^{T} \alpha_i} \quad (4)$$

We use this model for Platt Scaling and Isotonic Regression, and treat $f(x)$ as the raw (uncalibrated) prediction.

Platt [1999] uses a sigmoid to map SVM outputs on $[-\infty, +\infty]$ to posterior probabilities on $[0, 1]$. The sigmoidal shape of the reliability diagrams in Figure 2 suggest that the same calibration method should be effective at calibrating boosting predictions. In this section we closely follow the description of Platt's method in [Platt, 1999].

Let the output of boosting be $f(x)$ given by equation 4. To get calibrated probabilities, pass the output of boosting through a sigmoid:

$$P(y = 1|f) = \frac{1}{1 + exp(Af + B)} \quad (5)$$

where the $A$ and $B$ are fitted using maximum likelihood estimation from a calibration set $(f_i, y_i)$. Gradient descent is used to find $A$ and $B$ that are the solution to:

$$\underset{A,B}{argmin}\{-\sum_{i} y_i log(p_i) + (1 - y_i)log(1 - p_i)\}, \quad (6)$$

where

$$p_i = \frac{1}{1 + exp(Af_i + B)} \quad (7)$$

Two questions arise: where does the sigmoid training set $(f_i, y_i)$ come from and how to avoid overfitting to it.

One answer to the first question is to use the same train set as boosting: for each example $(x_i, y_i)$ in the boosting train set, use $(f(x_i), y_i)$ as a training example for the sigmoid. Unfortunately, this can introduce unwanted bias in the sigmoid training set and can lead to poor results [Platt, 1999].

An alternate solution is to split the data into a training and a validation set. After boosting is trained on the training set, the predictions on the validation set are used to fit the sigmoid. Cross validation can be used to allow both boosting and the sigmoid to be trained on the full data set. The data is split into C parts. For each fold one part is held aside for use as an independent calibration validation set while boosting is performed using the other C-1 parts. The union of the C validation sets is then used to fit the sigmoid parameters. Following Platt, experiments in this paper use 3-fold cross-validation to estimate the sigmoid parameters.

As for the second question, an out-of-sample model is used to avoid overfitting to the calibration set. If there are $N_+$ positive examples and $N_-$ negative examples in the calibration set, for each example Platt Calibration uses target values $y_+$ and $y_-$ (instead of 1 and 0, respectively), where

$$y_+ = \frac{N_+ + 1}{N_+ + 2}; \; y_- = \frac{1}{N_- + 2} \qquad (8)$$

For a more detailed treatment, and a justification of these target values see [Platt, 1999]. The middle row of Figure 2 shows the sigmoids fitted with Platt Scaling.

### 3.3 Isotonic Regression

An alternative to Platt Scaling is to use Isotonic Regression [Robertson et al., 1988]. Zadrozny and Elkan [2002; 2001] successfully used Isotonic Regression to calibrate predictions from SVMs, Naive Bayes, boosted Naive Bayes, and decision trees. Isotonic Regression assumes only that:

$$y_i = m(f_i) + \epsilon_i \qquad (9)$$

where $m$ is an isotonic (monotonically increasing) function. Given a training set $(f_i, y_i)$, the Isotonic Regression problem is finding the isotonic function $\hat{m}$ such that:

$$\hat{m} = \underset{z}{argmin} \sum (y_i - z(f_i))^2 \qquad (10)$$

A piecewise constant solution $\hat{m}$ can be found in linear time by using the pair-adjacent violators (PAV) algorithm [Ayer et al., 1955] presented in Table 1.

As in the case of Platt calibration, if we use the boosting train set $(x_i, y_i)$ to get the train set $(f(x_i), y_i)$ for Isotonic Regression, we introduce unwanted bias – in the fully separable case, boosting will order all the negative examples before the positive examples, so Isotonic Regression will output just a 0,1 function. An unbiased calibration set can be obtained using the methods discussed in Section 3.2. For the Isotonic Regression experiments we use the same 3-fold CV methodology used with Platt Scaling. The bottom row of Figure 2 shows functions fitted with Isotonic Regression.

Table 1: PAV Algorithm

| **Algorithm 1.** PAV algorithm for estimating posterior probabilities from uncalibrated model predictions. |
|---|
| 1   Input: training set $(f_i, y_i)$ sorted according to $f_i$ |
| 2   Initialize $\hat{m}_{i,i} = y_i$, $w_{i,i} = 1$ |
| 3   While $\exists \; i \; s.t. \; \hat{m}_{k,i-1} \geq \hat{m}_{i,l}$ <br>       Set $w_{k,l} = w_{k,i-1} + w_{i,l}$ <br>       Set $\hat{m}_{k,l} = (w_{k,i-1}\hat{m}_{k,i-1} + w_{i,l}\hat{m}_{i,l})/w_{k,l}$ <br>       Replace $\hat{m}_{k,i-1}$ and $\hat{m}_{i,l}$ with $\hat{m}_{k,l}$ |
| 4   Output the stepwise constant function: <br>       $\hat{m}(f) = \hat{m}_{i,j}$, for $f_i < f \leq f_j$ |

## 4 Empirical Results

In this section we apply the three scaling methods to predictions from boosted trees and boosted stumps on eight binary classification problems. The ADULT, COV_TYPE and LETTER problems are from the UCI Repository [Blake and Merz, 1998]. COV_TYPE was converted to a binary problem by treating the largest class as positives and the rest as negatives. LETTER was converted to boolean two ways. LETTER.p1 treats the letter "O" as positive and the remaining letters as negative, yielding a very unbalanced problem. LETTER.p2 uses letters A-M as positives and N-Z as negatives, yielding a well balanced problem. HS is the IndianPine92 data set [Gualtieri et al., 1999] where the class Soybean-mintill is the positive class. SLAC is a particle physics problem from collaborators at the Stanford Linear Accelerator, and MEDIS and MG are medical data sets.

First we use boosted stumps to examine the case when the data is not separable in the span of the base learners. We boost five different types of stumps by using all of the splitting criteria in Buntine's IND package [1991]. Because boosting can overfit [Rosset et al., 2004; Friedman et al., 2000], and because many iterations of boosting can make calibration worse (see Figure 1), we consider boosted stumps after 2,4,8,16,32,64,128,256,512,1024,2048,4096 and 8192 steps of boosting. The left side of Table 2 shows the cross-entropy on large final test sets for the best boosted stump models before and after calibration[1]. The results show that the performance of boosted stumps is dramatically improved by calibration or by optimizing to log-loss. On average calibration reduces cross-entropy by about 23% and squared error by about 14% (see Figure 6). The three post-training calibration methods (PLATT, ISO, and LO-

---
[1] To protect against an infinite cross-entropy we prevent the models from predicting exactly 0 or 1

Table 2: Cross-entropy performance of boosted decision stumps and boosted decision trees before and after calibration. Qualitatively similar results are obtained for squared error.

| PROBLEM | BOOSTED STUMPS | | | | | BOOSTED TREES | | | | |
|---|---|---|---|---|---|---|---|---|---|---|
| | RAW | PLATT | ISO | LOGIST | LOGLOSS | RAW | PLATT | ISO | LOGIST | LOGLOSS |
| COV_TYPE | 0.7571 | 0.6705 | 0.6714 | 0.6785 | 0.6932 | 0.6241 | 0.5113 | 0.5179 | 0.7787 | 0.7062 |
| ADULT | 0.5891 | 0.4570 | 0.4632 | 0.4585 | 0.4905 | 0.5439 | 0.4957 | 0.5072 | 0.5330 | 0.5094 |
| LETTER.P1 | 0.2825 | 0.0797 | 0.0837 | 0.0820 | 0.0869 | 0.0975 | 0.0375 | 0.0378 | 0.1172 | 0.0799 |
| LETTER.P2 | 0.8018 | 0.5437 | 0.5485 | 0.5452 | 0.5789 | 0.3901 | 0.1451 | 0.1412 | 0.5255 | 0.3461 |
| MEDIS | 0.4573 | 0.3829 | 0.3754 | 0.3836 | 0.4411 | 0.4757 | 0.3877 | 0.3885 | 0.5596 | 0.5462 |
| SLAC | 0.8698 | 0.8304 | 0.8351 | 0.8114 | 0.8435 | 0.8270 | 0.7888 | 0.7786 | 0.8626 | 0.8450 |
| HS | 0.6110 | 0.3351 | 0.3365 | 0.3824 | 0.3407 | 0.4084 | 0.2429 | 0.2501 | 0.5988 | 0.4780 |
| MG | 0.4868 | 0.4107 | 0.4125 | 0.4096 | 0.4568 | 0.5021 | 0.4286 | 0.4297 | 0.5026 | 0.4830 |
| MEAN | 0.6069 | 0.4637 | 0.4658 | 0.4689 | 0.4914 | 0.4836 | 0.3797 | 0.3814 | 0.5597 | 0.4992 |

GIST in the table) work equally well. Logistic Correction has the advantage that no extra time is required to fit the calibration model and no cross-validation is needed to create independent test sets. The LOGLOSS column shows the performance of boosted stumps when trained to optimize the log-loss as opposed to the usual exponential loss. When directly optimizing log-loss, the performance of boosted stumps is a few percent worse than with the other calibration methods. This is consistent with the results reported in [Lebanon and Lafferty, 2001]. As with Logistic Correction, there is no need for an independent calibration set and no extra time is spent training the calibration models.

Logistic Correction and directly optimizing log-loss are effective when using very weak base learners such as 1-level stumps. Unfortunately, because the base learners are so simple, boosted stumps are not able to capture all the structure of most real data-sets. Boosted 1-level stumps are optimal when the true model is additive in the features, but can not model non-additive interactions between 2 or more features [Friedman et al., 2000]. As we see in Table 2, boosting full decision trees yields significantly better performance on 5 of the 8 test problems.

Unlike 1-level stumps, decision trees are complex enough base models to make many data-sets separable (in their span). This means that boosted decision trees are expressive enough to capture the full complexity of most data-sets. Unfortunately this means they are expressive enough to perfectly separate the training set, pushing the probabilities predicted by Logistic Correction to 0 or 1. Although Logistic Correction is no longer effective, Figure 2 shows that good posterior estimates can still be found by fitting a sigmoid or an isotonic function on an *independent test set*.

The right side of Table 2 presents the performance of the best boosted trees before and after calibration. To prevent the results from depending on one specific tree style, we boost ten different styles of trees. We use all the tree types in Buntine's IND package [1991] (ID3, CART, CART0, C4.5, MML, SMML, BAYES) as well as three new tree types that should predict better probabilities: unpruned C4.5 with Laplacian smoothing [Provost and Domingos,

2003]; unpruned C4.5 with Bayesian smoothing; and MML trees with Laplacian smoothing. We consider the boosted models after 2,4,8,16,32,64,128,256,512,1024 and 2048 steps of boosting.

As expected, Logistic Correction is not competitive when boosting full decision trees. The other two calibration methods (Platt Scaling and Isotonic Regression) provide a significant improvement in the quality of the predicted probabilities. Both methods reduce cross-entropy about 21% and squared error about 13% (see Figure 6).

Surprisingly, when boosting directly optimizes log-loss, the performance of boosted trees is very poor. Because the base-level models are so expressive, boosting with log-loss quickly separates the two classes on the train set and pushes predictions toward 0 and 1, hurting calibration. This happens despite the fact that we regularize boosting by varying the number of iterations of boosting [Rosset et al., 2004].

Comparing the results in the left and right side of Tables 2 we see that boosted trees significantly outperform boosted stumps on five of the eight problems. On average over the eight problems, boosted trees yield about 13% lower squared error and 18% lower cross-entropy than boosted stumps. Boosted stumps, however, do win by a small margin on three problems, and have the nice property that the theoretically suggested Logistic Correction works well.[2]

To determine how the performance of Platt Scaling and Isotonic Regression changes with the amount of data available for calibration, we vary the size of the calibration set from 32 to 8192 by factors of two. Figure 5 shows the average squared error over the eight test problems for the best uncalibrated and the best calibrated boosted trees. We perform ten trials for each problem.

The nearly horizontal line in the graph show the squared error prior to calibration. This line is not perfectly horizon-tal only because the test sets change as more data is moved into the calibration sets. The plot shows the squared error after calibration with Platt's method or Isotonic Regression as the size of the calibration set varies from small to large. The learning curves show that the performance of boosted trees is improved by calibration even for very small calibration set sizes. When the calibration set is small (less than about 2000 cases), Platt Scaling outperforms Isotonic Regression. This happens because Isotonic Regression is less constrained than Platt Scaling, so it is easier for it to overfit when the calibration set is small. Platt's method also has some overfitting control built in (see Section 3).

---
[2] We also tried boosting 2-level decision trees. Boosted 2-level trees outperformed boosted 1-level stumps, but did not perform as well as boosting full trees. Moreover, 2-level trees are complex enough base-level models that Logistic Correction is no longer as effective as Platt Scaling or Isotonic Regression.

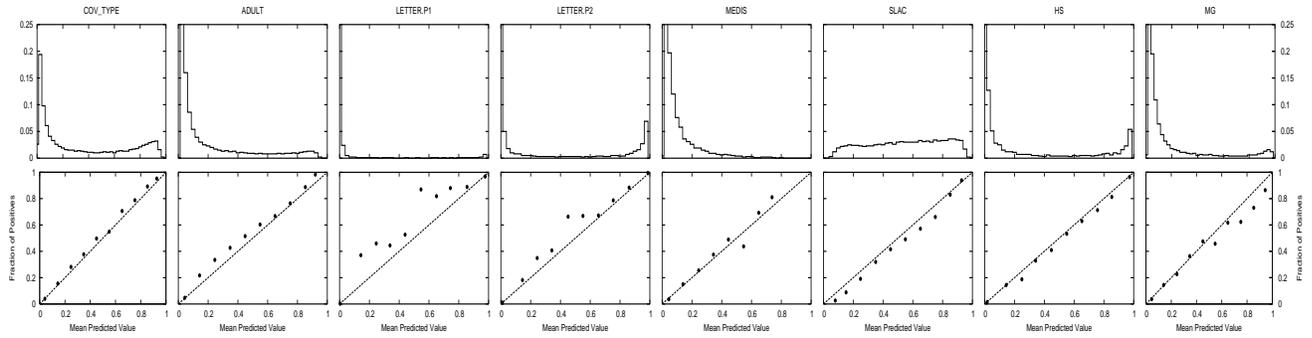

Figure 3: Histograms of predicted values and reliability diagrams for boosted trees calibrated with Platt's method.

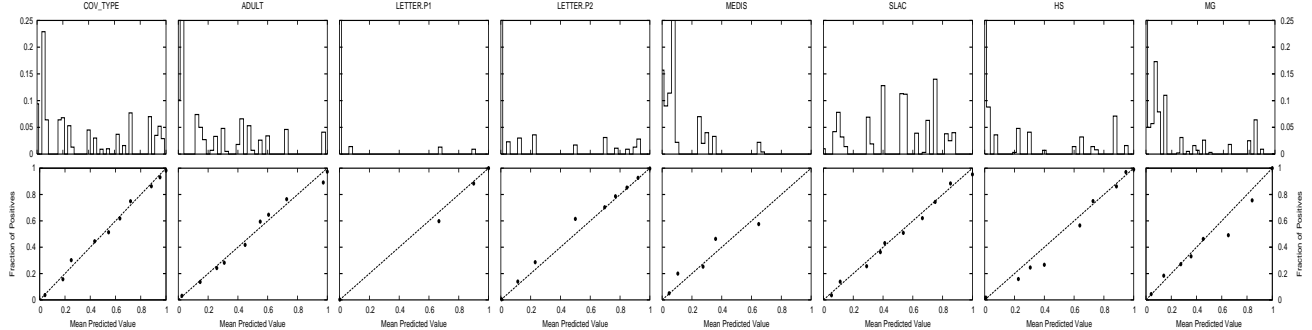

Figure 4: Histograms of predicted values and reliability diagrams for boosted trees calibrated with Isotonic Regression.

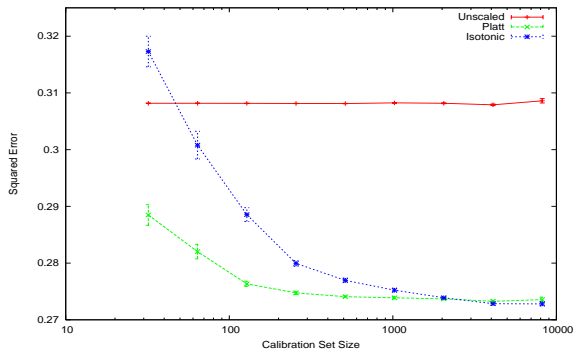

Figure 5: Learning curves for Platt Scaling and Isotonic Regression for boosted trees (average across 8 problems)

To illustrate how calibration transforms the predictions, we show histograms and reliability diagrams for the eight problems for boosted trees after 1024 steps of boosting, after Platt Scaling (Figure 3) and after Isotonic Regression (Figure 4). The figures show that calibration undoes the shift in probability mass caused by boosting: after calibration many more cases have predicted probabilities near 0 and 1. The reliability plots are closer to the diagonal, and the S shape characteristic of boosting's predictions is gone. On each problem, transforming the predictions using either Platt Scaling or Isotonic Regression yields a significant improvement in the quality of the predicted probabilities, leading to much lower squared error and cross-entropy. The main difference between Isotonic Regression and Platt Scaling for boosting can be seen when comparing the histograms in the two figures. Because Isotonic Regression generates a piecewise constant function, the histograms are quite coarse, while the histograms generated by Platt Scaling are smooth and easier to interpret.

## 5 Boosted Trees vs. Other Methods

Boosted trees have such poor initial calibration that it is not surprising that Platt Calibration and Isotonic Regression significantly improves their squared error and cross-entropy. But this does not necessarily mean that after calibration boosted trees will yield good probabilistic predictions. Figure 6 compares the performance of boosted trees and stumps with eight other learning methods before and after calibration. The figure shows the performance for boosted decision trees (BST-DT), SVMs, random forests

(RF), bagged decision trees (BAG-DT), neural nets (ANN), memory based learning methods (KNN), boosted stumps (BST-STMP), vanilla decision trees (DT), logistic regression (LOGREG), and naive bayes (NB). The parameters for each learning method were carefully optimized to insure a fair comparison.[3] To obtain the performance for pre-calibrated SVMs, we scaled SVM predictions to [0,1] by $(x - min)/(max - min)$.[4] Because of computational issues, all the models are calibrated on a held out 1K validation set instead of performing the 3-fold CV used until now. For each problem and each calibration method, we select the best model trained with each learning algorithm using the same 1K validation set used for calibration, and report it's performance on the final test set.

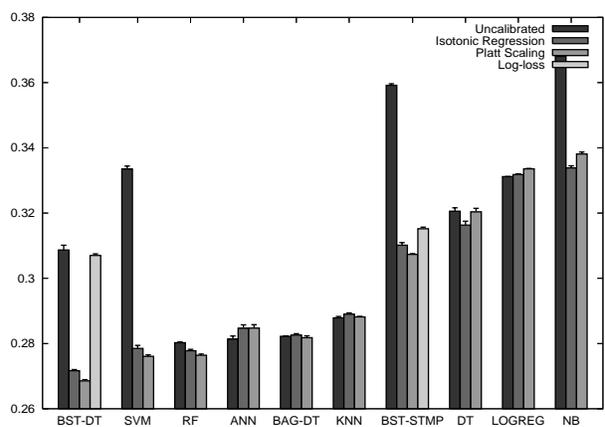

Figure 6: Squared error performance of the ten learning algorithms (on large test sets) before and after calibration. A qualitatively similar figure is obtained if one examines cross-entropy instead of squared error. Error bars are shown, but the confidence intervals are small so they may be difficult to see. The performance of boosting to optimize log-loss is shown for boosted trees and stumps.

After calibration with Platt Scaling or Isotonic Regression, boosted decision trees have better squared error and cross-entropy than the other nine learning methods. After calibration, boosting full decision trees can yield state-of-the-art probabilistic prediction. Boosting full trees to optimize log-loss directly, however, does not yield performance comparable to post-training calibration with either Platt Scaling or Isotonic Regression. Boosting to optimize log-loss directly is nearly as effective as calibration when boosting stumps, but boosted stumps perform much worse than boosting full decision trees.

After boosted decision trees, the next best methods are SVMs, random forests, neural nets, and bagged decision trees. While Platt Scaling and Isotonic Regression significantly improve the performance of the SVM models, and yield a small improvement with random forests, they have little effect on the performance of bagged trees and neural nets. Both bagged trees and neural nets yield well calibrated predictions that do not benefit from post-calibration. In fact, post-calibration hurts the performance of neural nets slightly. It is interesting to note that neural networks with a single sigmoid output unit can be viewed as a linear classifier (in the span of it's hidden units) with a sigmoid at the output that calibrates the predictions. In this respect neural nets are similar to SVMs and boosted decision trees *after* they have been calibrated using Platt's method.

---

[3]See [Caruana and Niculescu-Mizil, 2005] for a description of the different models and parameter settings that were evaluated.

[4]This is not a good way to scale SVM predictions. Platt Scaling and Isotonic Regression are much better. We use this simple scaling only to provide a baseline performance for SVMs.

Platt Scaling may be slightly less effective than Isotonic Regression with bagged trees, neural nets, memory based learning, vanilla decision trees, logistic regression, and naive bayes. This probably is because a sigmoid is not the correct function for mapping the raw predictions generated by these learning algorithms to posterior probabilities. In these cases, Isotonic Regression yields somewhat better squared error and cross-entropy than Platt Scaling, but still not enough to make these learning methods perform as well as calibrated boosted decision trees. The four learning methods that benefit the most from post-training calibration are boosted trees, SVMs, boosted stumps, and naive bayes. Calibration is much less important with the other learning methods. The models that predict the best probabilities prior to calibration are random forests, neural nets, and bagged trees. For a more detailed discussion on predicting well calibrated probabilities with supervised learning see [Niculescu-Mizil and Caruana, 2005].

## 6 Conclusions

In this paper we empirically demonstrated why AdaBoost yields poorly calibrated predictions. To correct this problem, we experimented with using a variant of boosting that directly optimizes log-loss, as well as with three post-training calibration methods: Logistic Correction justified by Friedman et al.'s analysis, Platt Scaling justified by the sigmoidal shape of the reliability plots, and Isotonic Regression, a general non-parametric calibration method.

One surprising result is that boosting with log-loss instead of the usual exponential loss works well when the base level models are weak models such as 1-level stumps, but is not competitive when boosting more powerful models such as full decision trees. Similarly Logistic Correction works well for boosted stumps, but gives poor results when boosting full decision trees (or even 2-level trees).

The other two calibration methods, Platt Scaling and Isotonic Regression, are very effective at mapping boosting's predictions to calibrated posterior probabilities, regardless of how complex the base level models are. After calibration with either of these two methods, boosted decision trees have better reliability diagrams, and significantly improved squared-error and cross-entropy.

When compared to other learning algorithms, boosted decision trees that have been calibrated with Platt Scaling or Isotonic Regression yield the best average squared error and cross-entropy performance. This, combined with their excellent performance on other measures such as accuracy, precision, and area under the ROC curve [Caruana and Niculescu-Mizil, 2005], may make calibrated boosted decision trees the model of choice for many applications.

**Acknowledgments**

Thanks to Bianca Zadrozny and Charles Elkan for use of their Isotonic Regression code. Thanks to Charles Young et al. at SLAC (Stanford Linear Accelerator) for the SLAC data set. Thanks to Tony Gualtieri at Goddard Space Center for help with the Indian Pines Data set. This work was supported by NSF Award 0412930.